\documentclass[journal]{IEEEtran}
\IEEEoverridecommandlockouts
\usepackage{cite}
\usepackage{amsmath,amssymb,amsfonts}
\usepackage{algorithmic}
\usepackage{graphicx}
\usepackage{textcomp}
\usepackage[table,xcdraw]{xcolor}
\usepackage{url}
\usepackage{multirow}
\usepackage{booktabs}
\usepackage{balance}
\usepackage{caption}
\usepackage{subcaption}
\usepackage{balance}

\begin{document}
\title{SegCodeNet: Color-Coded Segmentation Masks for Activity Detection from Wearable Cameras}

\author{
        Asif Shahriyar Sushmit$^{1}$, 
        Partho Ghosh$^{2}$,
        Md.Abrar Istiak$^{2}$,
        Nayeeb Rashid$^{2}$,
        Ahsan Habib Akash$^{2}$,
        \\Taufiq Hasan$^{1}$$^{\dagger}$
        \thanks{$^{1}$mHealth Research Group, Department of Biomedical Engineering (BME), Bangladesh University of Engineering and Technology (BUET), Dhaka - 1205, Bangladesh. Email: {\tt\scriptsize sushmit@ieee.org, taufiq@bme.buet.ac.bd}}
        \thanks{$^{2}$Department of Electrical and Electronics Engineering (EEE), Bangladesh University of Engineering and Technology, Dhaka - 1205, Bangladesh }
        \thanks{$^{\dagger}$This work was supported by the Department of BME, BUET and Brain Station 23 (Dhaka - 1212, Bangladesh). The TITAN Xp GPU used for this work was donated by the NVIDIA Corporation.}
 }

\maketitle
\thispagestyle{plain}
\pagestyle{plain}

\begin{abstract}
Activity detection from first-person videos (FPV) captured using a wearable camera is an active research field with potential applications in many sectors, including healthcare, law enforcement, and rehabilitation. State-of-the-art methods use optical flow-based hybrid techniques that rely on features derived from the motion of objects from consecutive frames. In this work, we developed a two-stream network, the \emph{SegCodeNet}, that uses a network branch containing video-streams with color-coded semantic segmentation masks of relevant objects in addition to the original RGB video-stream. We also include a stream-wise attention gating that prioritizes between the two streams and a frame-wise attention module that prioritizes the video frames that contain relevant features. Experiments are conducted on an FPV dataset containing $18$ activity classes in office environments. In comparison to a single-stream network, the proposed two-stream method achieves an absolute improvement of $14.366\%$ and $10.324\%$ for averaged F1 score and accuracy, respectively, when average results are compared for three different frame sizes $224\times224$, $112\times112$, and $64\times64$. The proposed method provides significant performance gains for lower-resolution images with absolute improvements of $17\%$ and $26\%$ in F1 score for input dimensions of $112\times112$ and $64\times64$, respectively. The best performance is achieved for a frame size of $224\times224$ yielding an F1 score and accuracy of $90.176\%$ and $90.799\%$ which outperforms the state-of-the-art Inflated 3D ConvNet (I3D) \cite{carreira2017quo} method by an absolute margin of $4.529\%$ and $2.419\%$, respectively.
\end{abstract}

\begin{IEEEkeywords}
Activity classification, segmentation mask, first-person video.
\end{IEEEkeywords}

\section{Introduction}
The wide-spread availability of body-worn cameras enables a multitude of new applications in daily life using first-person videos. With the large quantity of video data becoming available, automatic processing of FPVs has become a topic of greater interest for human activity recognition (HAR). Various application domains of HAR include activity logs, law enforcement, healthcare, search and rescue missions, inspections, home-based rehabilitation, sporting activity observation \cite{poleg_wacv16_compactcnn} and wildlife observation \cite{tadesse2018visual}. In the wake of the COVID-19 pandemic, activity recognition from videos can become vital in detecting if the users are adhering to social distancing and hand hygiene guidelines. These applications can be of significance once the governments are beginning to resume regular activities.

A vast amount of research work has already been done in the area of HAR. The research topic of HAR can be broadly classified into three different streams \cite{hussain2019different}: (i) radio frequency-based, (ii) sensor-based \cite{paulson2011object}, and (iii) video-based. The focus of the current work is on videos, which can again be categorized into two types, namely third-person video (TPV) and first-person video (e.g., recorded by a wearable camera). Some of the most popular datasets on available for human activity recognition are UCF101 \cite{soomro2012ucf101}, HMDB-51 \cite{Kuehne11}, THUMOS \cite{idrees2017thumos} and the Kinetics datasets  \cite{kay2017kinetics}, \cite{carreira2018short} and \cite{carreira2019short}, all of which mostly contain third-person videos of people performing different tasks. The first publicly available dataset consisting of FPV for activity recognition was collected in a controlled office setting \cite{mayol2005wearable}. Other available FPV datasets include CMU-MMAC \cite{spriggs2009temporal}, GTEA-11\cite{fathi2011understanding}, VNIST \cite{aghazadeh2011novelty}, HUJI EgoSeg \cite{poleg_wacv16_compactcnn,poleg_cvpr14_egoseg}. These datasets offer videos from the head-, shoulder-, and chest-mounted wearable cameras. For both FPV and TPV domains, the major challenge in automatic HAR includes scale and texture variation, low-resolution, motion-blur, illumination changes, context analysis, and self-occlusions \cite{abebe2016robust}, \cite{abebe2017hierarchical}. FPV, on the other hand, poses some additional challenges due to a more dynamic background, scarcity of information, and unstable perspective. 

\begin{figure}[t]
\centering
\includegraphics[width=\linewidth]{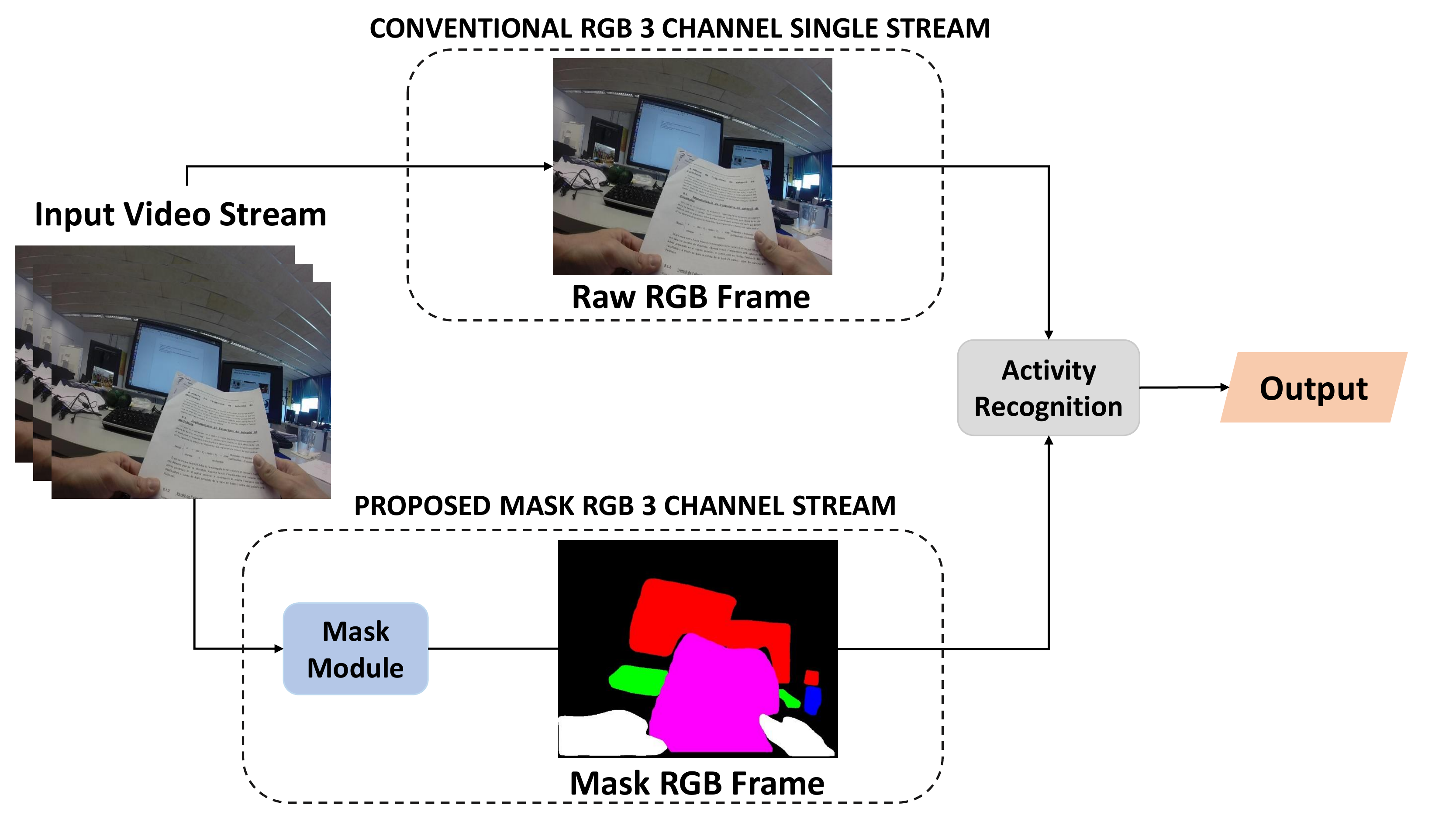}
\caption{Overview of the proposed two-stream network architecture showing the conventional RGB-stream and the proposed segmentation mask based stream.}
\end{figure}

The existing literature on activity recognition using image and video processing techniques can be broadly classified into methods based on traditional features \cite{peng2016bag,wang2013action,oneata2013action,peng2014action}, and neural networks. Traditional methods mainly focus on designing useful features to be extracted from the video frames to be classified by machine learning methods. The unimodal (i.e., single-stream) methods for human activity classification can be divided into four categories \cite{vrigkas2015review}: space-time methods, rule-based methods, shape-based methods, and stochastic methods. 

Methods involving a neural network can be further categorized into four groups that are based on: (i) single RGB stream videos, (ii) optical-flow, (iii) pose estimation, and (iv) a hybrid approach. RGB stream based techniques \cite{ullah2017action,ludl2019simple} generally include a feature extractor that inputs the original video frame pixels and a recurrent neural network for classification. Popular feature extractors include variations of AlexNet \cite{krizhevsky2012imagenet}, ResNet \cite{he2016deep}, Wide ResNet \cite{zagoruyko2016wide}, ResNeXt \cite{xie2017aggregated}, DenseNet \cite{huang2017densely} and Xception \cite{chollet2017xception}. 
For the recurrent network used for classification, one or more uni/bi-directional LSTM (Long Short Term Memory) layers are used \cite{schuster1997bidirectional}. Pose estimation based methods determine the position and orientation of the different limbs of the human beings present in the fields of view \cite{ludl2019simple, cao2018openpose}. 
Hybrid methods \cite{hong2019contextual,li2018rehar} include methods that augment the feature extraction process by providing additional information via network branches. Notable methods include additional features extracted using eye and ego motion \cite{ogaki2012coupling}, pose estimation \cite{du2017rpan}, hand segmentation \cite{serra2013hand} and optical-flow \cite{horn1981determining, ilg2017flownet}. The optical flow based methods \cite{carreira2017quo, choutas2018potion} are currently the state of the art for the popular activity classification datasets such as UCF-101 \cite{soomro2012ucf101} and HMDB-51 \cite{Kuehne11}. 

Optical flow computation from successive video frames creates binary or colorful masks that attribute different contrast/color values for each of the pixels that are changing temporally. However, for non-stationary scenes, optical flow-based systems suffer from noisy features and classification errors due to a rapidly changing background. To address this problem, \cite{huang2019efficient} proposes a neural network-based approach that attempts to remove the noise-induced in the optical flow due to motion artifacts. However, a fundamental problem with the optical flow based methods is the inherent assumption that anything that is moving is important for activity classification, which may not always be correct.

In this work, we propose the SegCodeNet, a simple but effective two-stream approach that leverages the information extracted from task-relevant object segmentation masks. Previous research \cite{fathi2011understanding} shows that knowing the action improves object recognition performance. Conversely, we hypothesize that knowing the objects should help in classifying action. Following the hybrid methods \cite{hong2019contextual,li2018rehar}, the proposed SegCodeNet includes an additional branch in the network, including segmentation masks from objects relevant for classification. We first generate semantic segmentation masks from the input video frames using a Mask R-CNN network \cite{he2017mask}. Instead of providing a collapsed binary segmentation mask that is not effective for multiple objects \cite{hong2019contextual}, we propose a novel color-coded mask where unique colors are attributed to multiple task-relevant objects. This approach simultaneously retains information regarding the object's presence, boundary, and motion information that can be utilized by the subsequent feature extractor and classifier network. We expect our approach to be superior to optical flow-based techniques since only task-relevant objects are visible in the segmentation stream, which is not affected by the movements in the background. The proposed network thus exploits the interrelation between the objects present in the field of view and their contribution to activity recognition. The weighted and merged features from the masked and the RGB stream enable our model to discover detailed Spatio-temporal patterns with enriched semantic information. The presented system also incorporates stream-wise and frame-wise attention gates to ensure the prioritization of the most relevant features. 

This paper is organized as follows. Section II describes the dataset used for the study and discusses the challenges involved. In Sec. III, we describe the proposed two-stream architecture and its various modules in detail. Section IV mentions the two state-of-the-art baseline models used for comparison, followed by detailed experimental evaluations in Sec. V. Results are further discussed in Sec. VI before the paper is concluded in Sec. VII.

\begin{figure}[t]
\centering
\includegraphics[trim={0 10 75 10}, clip,width=\linewidth]{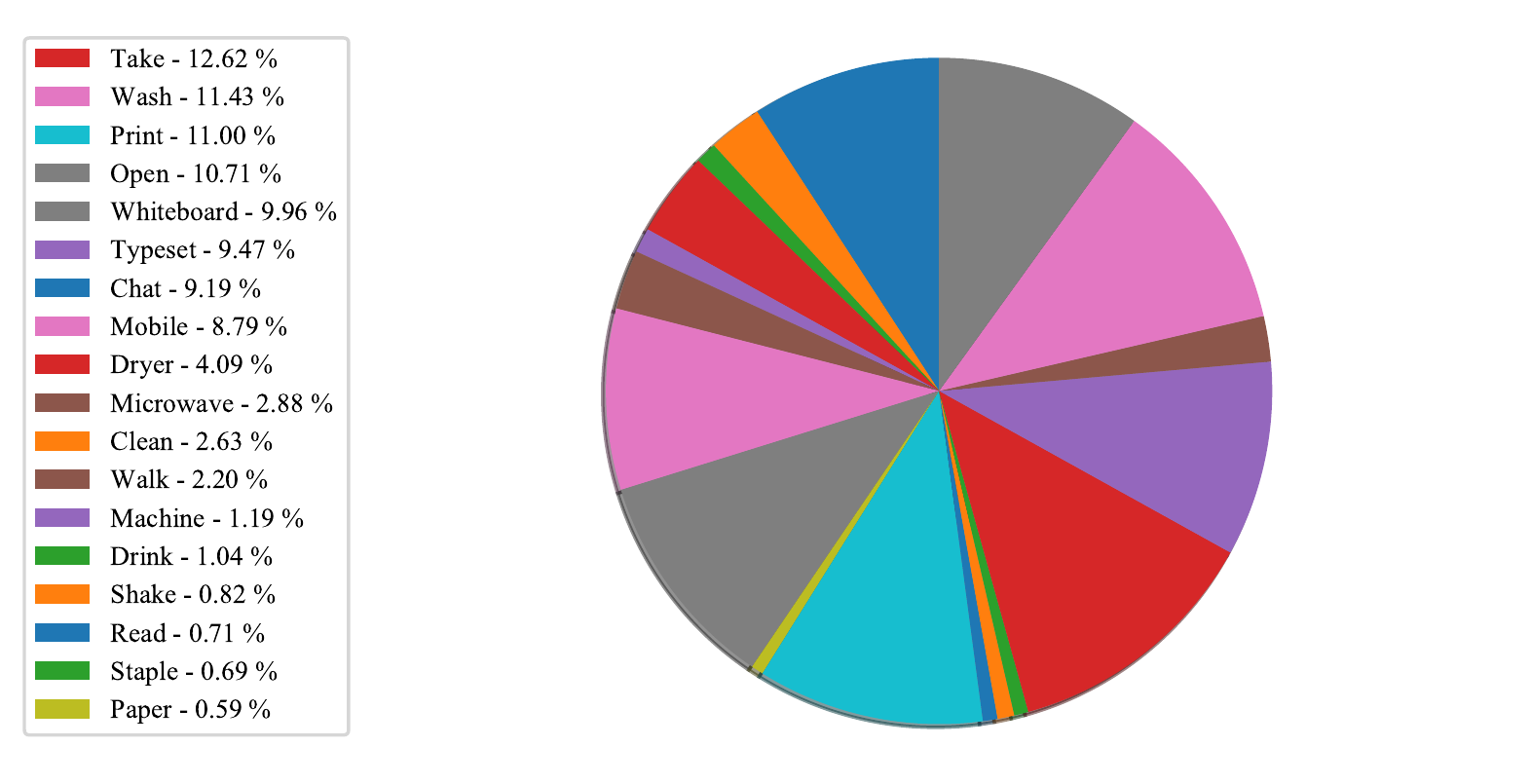} 
\caption{Distribution of various activity classes in the FPV-O dataset showing the class imbalance issue.}
\label{data_piechart}
\end{figure}

\begin{figure*}[t]
    \centering
    \includegraphics[width=\linewidth]{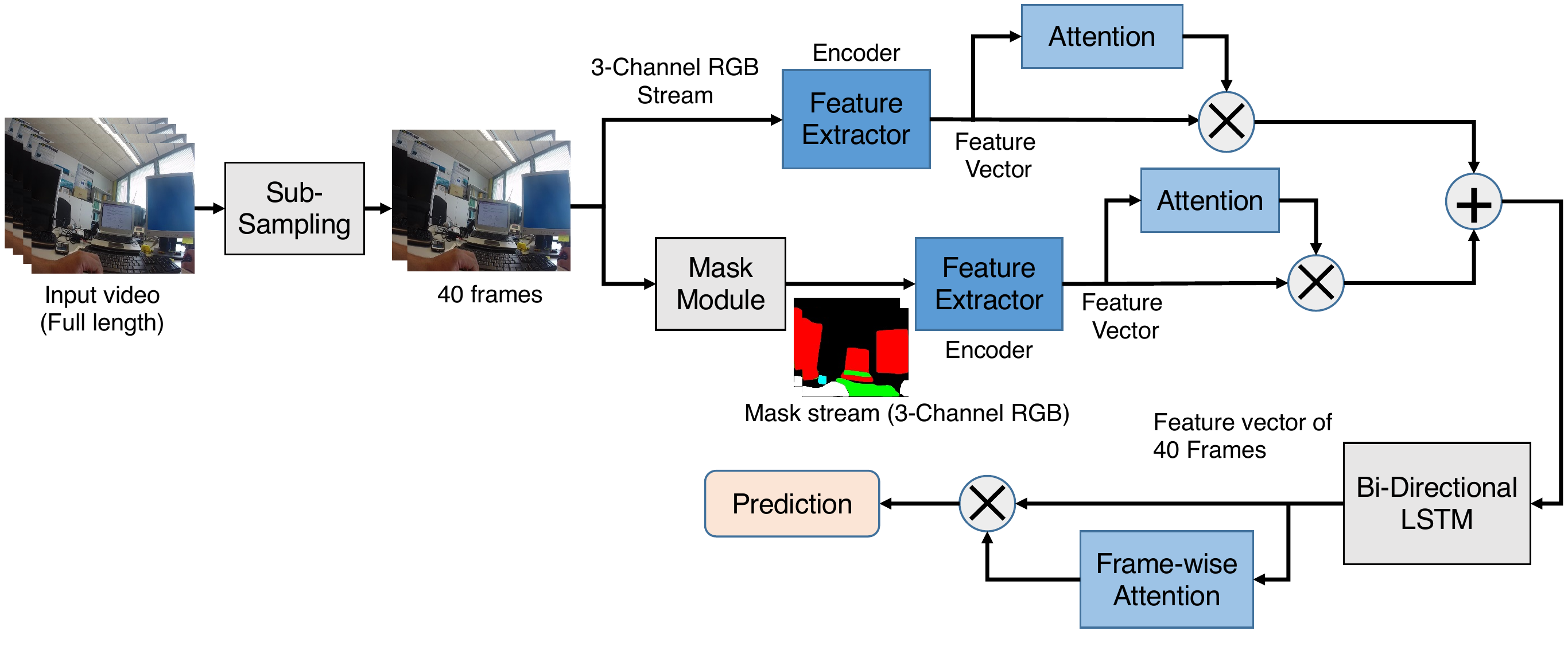}
    \vspace{-7mm}
    \caption{Overview for the proposed two-stream activity classification architecture. The network constructs two different feature vectors which are concatenated and are input to a bidirectional LSTM for activity classification. The Mask Module converts the RGB frames into segmented and color-coded masks based on detected relevant objects. The `$\mathbf{\times}$' signs denote element wise multiplication and the `$\mathbf{+}$' sign denotes concatenation of the incoming vectors.}
    \label{systemDiag}
\end{figure*}

\section{Dataset}
The dataset used in this work has been provided through the IEEE VIP Cup 2019 competition \cite{vip_cup_website,abebefirst} 
which consists of first-person videos in office settings. We refer to this dataset as FPV-O for the remaining of this paper. The videos in this dataset were collected using a chest-mounted GoPro Hero3+ Camera with a resolution of $1280\times760$ pixels and a $30$ fps frame rate \cite{abebefirst}. 
There are four types of human activities present in FPV-O data: (i) ambulatory motion, (ii) human to human interaction, (iii) human to object interaction, and (iv) solo activity. Overall, the dataset contains a total of $18$ activity classes. A small percentage of the videos in this dataset was recorded outdoors, where the lighting effect was noticeably different and had a frame-rate of $120$ fps. A summary of all the classes and the percentage of associated frames is presented in Fig. \ref{data_piechart}. From the figure, it is evident that significant class imbalance exists in the dataset, which needs to be addressed. The number of video segments in training and testing were $1230$ and $568$, respectively.



\section{Proposed Two-Stream Architecture}
The overall workflow of our architecture is presented in Fig. \ref{systemDiag}. Different blocks of the proposed activity classification system are described in the following sub-sections.

\subsection{Video sub-sampling}
First, we sub-sample the videos in order to reduce the computational load. The sub-sampling process extracts a fixed number of video frames from each video file irrespective of its length. The process described is as follows. 

Let the $n$-th frame of a video segment denoted by $\mathbf{V}[n]$, where $n \in [1, N]$. We first compute the average sampling period, $\tau = \lfloor N/k \rfloor$, where  $\lfloor \cdot \rfloor$ denotes the floor operation and $k$ denotes the fixed number of frames to extract. The first frame of the sub-sampled video is randomly selected as 
\begin{equation}
    i_{\mbox{\tiny rand}} = \mbox{Rand}(1,\tau)
\end{equation}
where Rand$(i,j)$ selects a random integer between $[i,j]$ (inclusive set). The subsequent frames are uniformly selected between an interval of $\tau$ frames. Therefore, the $i$-th frame of the sub-sampled video segment can be obtained as 
\begin{equation}
    \mathbf{\hat{V}}[i] = \mathbf{V}[i_{\mbox{\tiny rand}}+(i-1)\tau], \mbox{where $\{i | i \in \mathbb Z \cap [1, k]\}$}.
\end{equation}
This sub-sampling scheme ensures that the information content of the entire video is captured within a fixed number of frames. For the FPV-O dataset, we use $k=40$ as we have empirically found that this particular value performs better compared to $k = 20$ and $k \geq 60$. Another value of $k$ may be more suitable for a different datasets.

\subsection{Generating segmentation masks}\label{segmentation mask}
\begin{figure}[t]
    \centering
    \includegraphics[width=\linewidth]{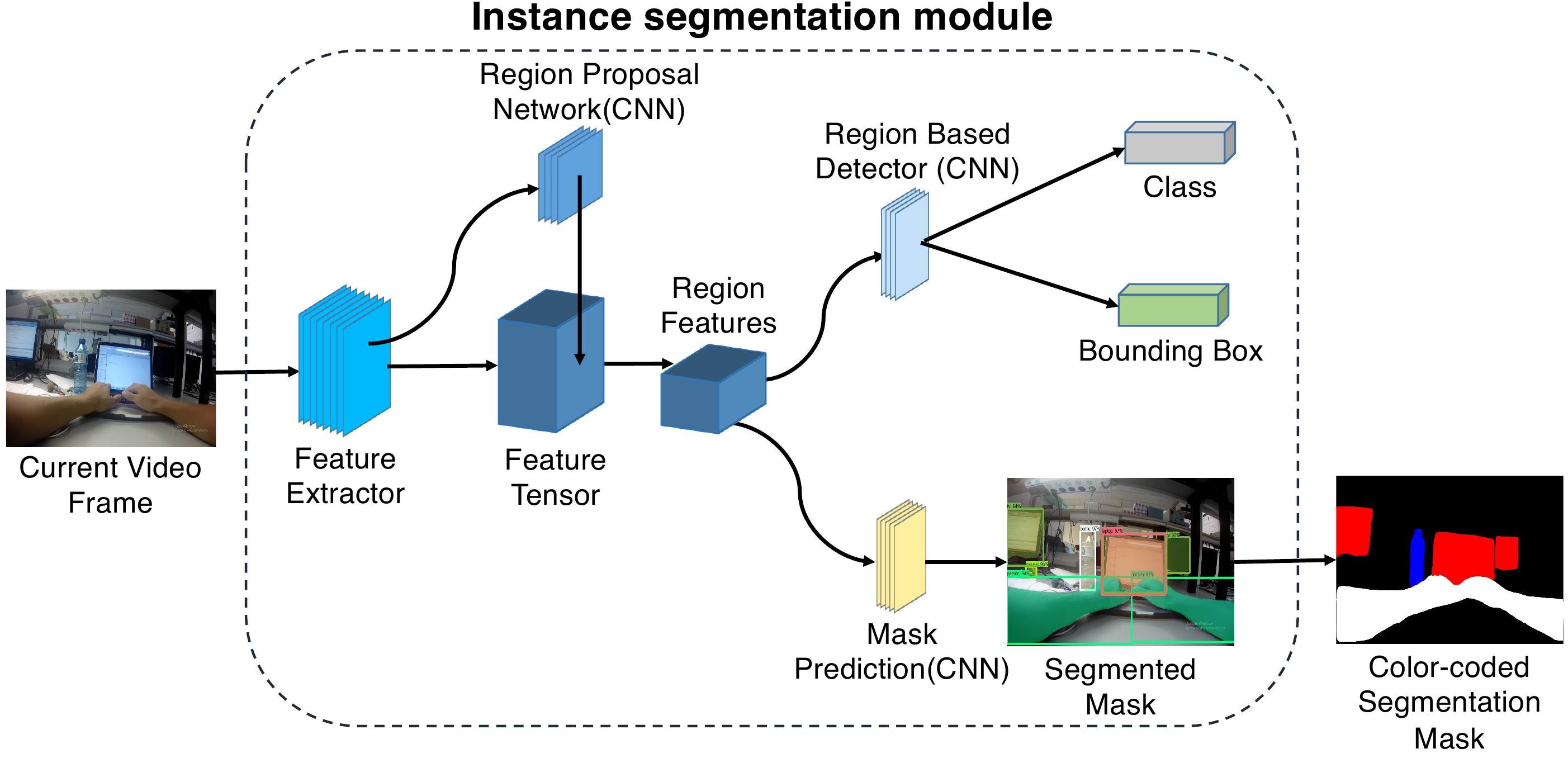}
    \caption{Proposed color coded mask generation scheme of SegCodeNet. A mask R-CNN architecture is used for instance segmentation while identified segmentation masks are color-coded according to Table \ref{table:colorscheme}.}
    \label{masknetwork}
\end{figure}

The mask-based stream in the proposed network acts as a surrogate of the conventional optical-flow based streams used in state-of-the-art systems \cite{carreira2017quo, choutas2018potion}. First, we manually identify important objects for each activity class in the FPV-O dataset. For instance, ``digital screen", ``laptop", ``paper", ``person" (hand), etc. objects are usually observed in the ``Read" activity class. Based on all the target activity classes, we identify an important set of objects and assigned a unique color to each of them, as mentioned in Table \ref{table:colorscheme}. These $14$ objects are a subset of the $80$ objects available in COCO dataset \cite{lin2014microsoft}.
To generate the masks, we first perform instance segmentation using the Mask R-CNN network \cite{he2017mask} from each video frame. Next, the class-relevant objects (Table \ref{table:colorscheme}) are filtered out from the segmented mask, followed by assigning the assigned color-code to obtain the final segmentation mask. The entire process is summarized in Fig. \ref{masknetwork}. 



Different networks pre-trained on the COCO dataset can be used as a feature extractor with a Mask R-CNN model. Popular networks include InceptionV2 \cite{szegedy2017inception},  ResNet50  \cite{he2016deep}, ResNet101 \cite{he2016deep}  and  Inception ResNet V2 \cite{szegedy2017inception} with  Atrous  convolution. In our system, we use the Inception ResNet V2 as it is known to provide superior segmentation performance compared to the other networks \cite{rosenfeld2018elephant}. 

Our fundamental assumption is that if these class-relevant objects are segmented and color-coded to generate a separate video-stream, the classifier can perform better with this additional information embedded within this video stream.

\begin{table}[t]
\centering
\caption{Coloring scheme for segmentation masks obtained from different class-relevant objects}
\label{table:colorscheme}
\begin{tabular}{lll}
\hline
\bf Relevant objects       & \bf Color  & \bf Decimal code (R,G,B)  \\ 
\hline
Person                &White & $(255,255,255)$ \\
TV, laptop, monitor & Red & $(255,0,0)$ \\
Bottle, cup, wine glass & Blue & $(0,0,255)$  \\
Cell phone & Cyan & $(0,255,255)$\\
Microwave, oven & Yellow &     $(255,255,0)$ \\
Sink & Light Blue & $(100,150,200)$ \\
Paper, book & Magenta & $(255,0,255)$ \\
Keyboard & Green & $(0,255,0)$\\
Background/irrelevant objects & Black & $(0,0,0)$\\
\hline
\end{tabular}
\end{table}

\subsection{Feature extractor}
In our proposed two-stream architecture, we use a ResNeXt-50 \cite{xie2017aggregated} feature extractor for both of the streams for activity classification in the next stage. Each sub-sampled frame is passed through the feature extractors in the two streams to obtain the corresponding feature vector. The feature vectors extracted from the $n$-th frame from the RGB and Mask-based streams are denoted by $\mathbf{F}_{RGB}[n]$ and $\mathbf{F}_{mask}[n]$, respectively.
We note that, the proposed architecture is not dependent on a specific feature extractor and alternative features could also have been used. 
Our primary motivation is that the original RGB video stream and the color-coded segmentation mask stream will provide complementary information that will eventually help the activity classifier in providing improved performance.

\subsection{Stream-wise attention}
The next stage of our architecture includes a stream-wise attention module that multiplies the extracted feature vectors $\mathbf{F}_{RGB}[n]$ and $\mathbf{F}_{mask}[n]$ by the learn-able scalar parameters $\mathbf{\eta}_{RGB} \in [0,1]$ and $\mathbf{\eta}_{mask} \in [0,1]$, respectively. The attention values for the RGB and masked streams are independently learned. We presume that for some activity classes the RGB video-stream contains more activity-relevant information compared to the masked stream, and vise versa. 


\makeatletter
\newcommand*\bigcdot{\mathpalette\bigcdot@{.5}}
\newcommand*\bigcdot@[2]{\mathbin{\vcenter{\hbox{\scalebox{#2}{$\m@th#1\bullet$}}}}}
\makeatother

\subsection{Bi-directional LSTM}
After the attention layer, the feature vectors extracted from the two streams are concatenated and passed to the Bi-directional LSTM layer \cite{schuster1997bidirectional} for temporal feature analysis. The feature vector received in this layer can be denoted as
\begin{equation}
    \mathbf{F}_{LSTM}[n]=\left({\eta}_{RGB} \bigcdot \mathbf{F}_{RGB}[n]\right) \mathbin\Vert \left({\eta}_{mask} \bigcdot \mathbf{F}_{mask}[n]\right)
\end{equation}
where $\mathbin\Vert$ is the matrix concatenation operator.
The bi-directional LSTM module consists of a single hidden layer. During our experiments, we observed that increasing the number of layers did not provide any significant advantage.


\subsection{Frame-wise attention}
Naturally, all the frames sampled from a video do not include frames containing relevant activity information. Thus, we also use a frame-wise attention module to provide a higher weight to the feature vectors corresponding to specific video frames from the sub-sampled data that are more important for classification.

\section{Baseline systems}
To evaluate the performance of our system, we have compared the proposed network with two state-of-the-art architectures for activity classification, including the I3D model \cite{carreira2017quo} and CNN feature-based Bi-directional LSTM model proposed by Amin Ullah \emph{et al.} \cite{ullah2017action}. 

\subsection{I3D baseline: Optical-flow hybrid network}
The I3D model \cite{carreira2017quo} is an optical flow-based state-of-the-art method achieving one of the top scores in the action classification datasets, HMDB-51 \cite{Kuehne11} and UCF101 \cite{soomro2012ucf101}. To implement the I3D model on the FPV-O dataset, we utilize the pre-trained weights obtained from the Kinetics dataset \cite{kay2017kinetics} and trained it with $64$ RGB and $64$ flow images (computed using the TV-L1 algorithm \cite{perez2013tv}) as described in \cite{carreira2017quo}. In the two-stream I3D architecture, a 3D CNN is used \cite{ji20123d}. In the 3D convolutional layer, an ImageNet pre-trained Inception-V1 model \cite{ioffe2015batch} is used as the base network with weights from the $N\times N$  filters were inflated to $N\times N \times N$ by repeating the weights of the 2D filters $N$ times along the time dimension, and dividing them by $N$ for re-scaling. This allows the model to capture both spatial and temporal information from the videos. 

\subsection{Single RGB stream method: CNN with Bi-directional LSTM}
This baseline follows \cite{ullah2017action} where the AlexNet \cite{krizhevsky2014one} feature extractor is used followed by used a Bi-Directional LSTM with two hidden layers. No attention layers were used. In this work, for the sake of comparison with the two-stream network, we implement this baseline system by using only the RGB stream part of the proposed network with the ResNeXt feature extractor. 

\section{Experimental evaluation}
\subsection{Implementation details}
We developed our model using Pytorch. The RAdam \cite{liu2019variance} optimizer was used for training. The segmentation masks were generated using a pre-trained Mask R-CNN \cite{he2017mask} model as described in Sec. \ref{segmentation mask}. 


Before feature extraction in the two-streams, the image frames obtained from the RGB and mask video streams were resized to $64\times64$, $112\times112$, or $224\times224$ depending on the experiment.
We used single-precision images to reduce the computational burden. 


The RGB and mask feature extractors were trained in two steps. In the first step, the pre-trained weights from Imagenet \cite{deng2009imagenet} and the weights were frozen and trained for $100$ epochs with a learning rate of $0.001$ (using corresponding RGB or Mask video data). In the second step, we load the best weights from the first phase, unfreeze the feature extractor and train for another $100$ epochs with a learning rate of $0.0001$. In these two steps of training, the number of mini-batches were selected differently for experiments with different image dimensions due to memory issues. Once the two streams are trained, all the parameters of the proposed architecture are trained on the dataset except for the feature extractors in the final stage. However, for video dimension $64\times64$ it was possible to train all the parameters in the final stage. The experiments were implemented on an NVIDIA Titan Xp Graphics processing unit (GPU). The codes for the proposed system available in our Github repository\footnote{https://github.com/mHealthBuet/SegCodeNet}.

\begin{figure}[t]
    \centering
    \includegraphics[trim={10 0 10 25},clip, width=\linewidth]{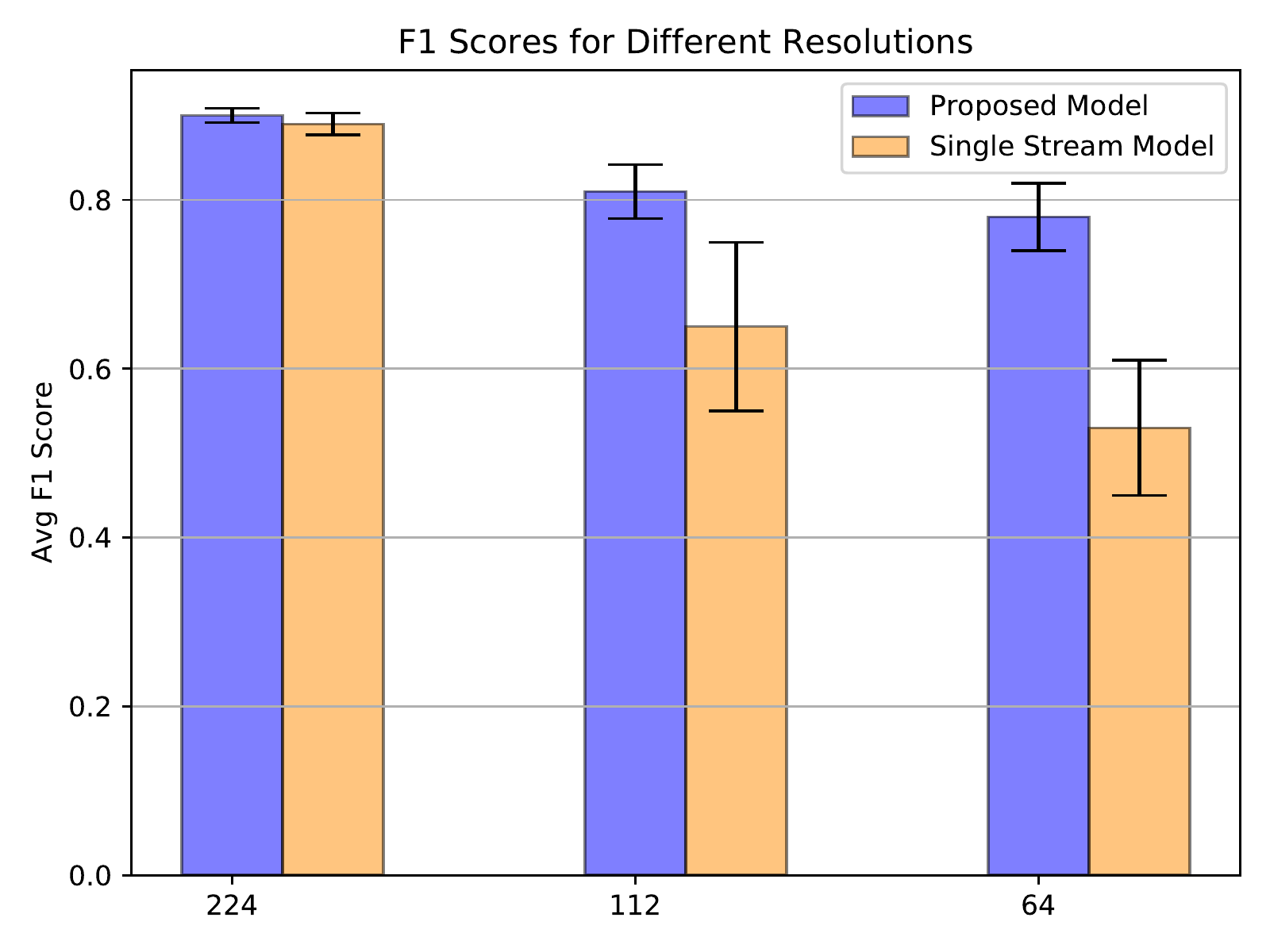}
    \caption{The averaged class-wise F1 scores and error-bars obtained while varying the input video resolution for the proposed two-stream and the baseline single-stream models. 
    } \label{errorbars}
\end{figure}

\begin{figure}[t]
    \includegraphics[trim={22 0 37 20},clip,width=\linewidth]{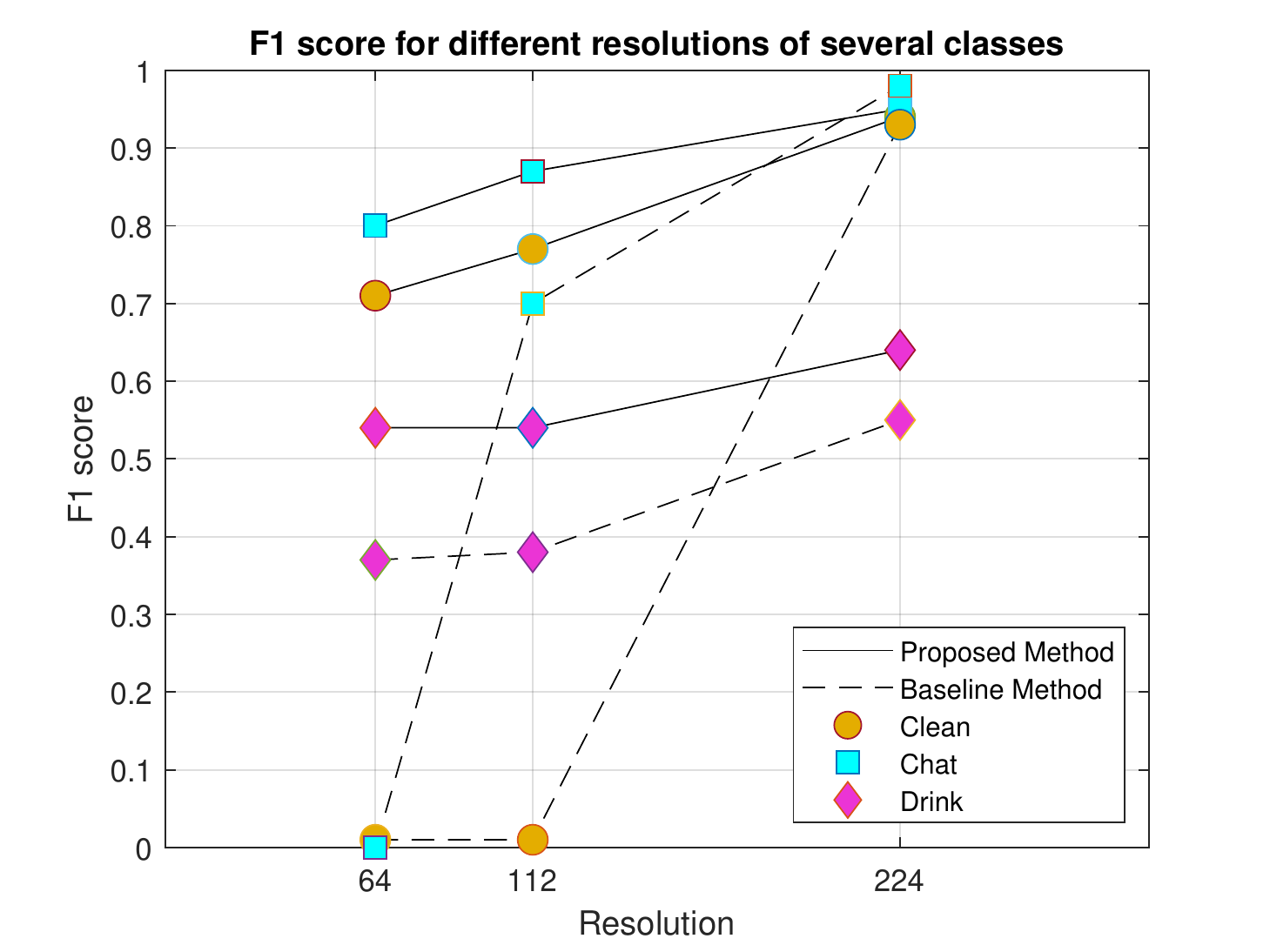}
    \caption{Comparison of the proposed system and the single-stream RGB baseline system in terms of averaged class-wise F1 scores on selected challenging activity classes (\emph{Clean}, \emph{Chat}, \emph{Drink}). Results are presented for different video frame sizes. 
    }
\label{clsf1}
\end{figure}
\begin{table}[b]
\caption{Comparison of the proposed two-stream architecture with the single-stream RGB baseline. Performance metrics are averaged across three different video dimensions.}
\label{baseline_average_sizes}
\centering
\begin{tabular}{lcc}
\toprule
Activity classification system & Avg F1 score (\%) & Accuracy (\%) \\ \midrule
Single-stream RGB system & 68.921 &  73.611\\
Proposed two-stream system & \bf 83.287 & \bf 83.935\\
\bottomrule
\end{tabular}
\end{table}
\subsection{Effectiveness of the mask stream}
First, we compared the overall performances of the proposed two-stream method with the single-stream RGB model \cite{ullah2017action}. In this experiment, we have used different video frame sizes $224\times224$, $112\times112$, and $64\times64$ and presented the averaged performance metrics for three different image resolutions. The averaged results obtained are presented in Table \ref{baseline_average_sizes}. These results show that the proposed method in comparison to the single-stream baseline provides with absolute gains of $14.366\%$ and $10.324\%$ with respect to averaged F1 score and accuracy, respectively. This demonstrates the effectiveness of using the segmentation mask stream that provides additional information to the activity classifier. 

\subsection{Effect of input video resolution}
To examine the effect of input video resolution, experiments were performed in three different video frame resolutions, $64\times64$, $112\times112$, and $224\times224$. The averaged class-wise F1 scores obtained from each of these video resolutions are presented in Fig. \ref{errorbars}. Here, we observe that the performance improvement achieved by the proposed method is more significant in the lower resolution images of dimension $64\times64$ and $112\times112$.
The percent absolute improvements in F1 scores achieved by the proposed system are $1\%$, $17\%$ and $26\%$ for video resolutions of $224\times224$, $112\times112$ and $64\times64$, respectively. We performed the McNemar's statistical significance test \cite{dietterich1998approximate} for these three experiments and found that the improvements are significant ($p < 0.01$) for the dimensions $64\times64$ and $112\times112$. We believe this to be a noteworthy achievement of the two-stream method. 

To gain a deeper understanding of why our method is able to provide such significant improvements in lower resolutions, we identified three difficult activity classes, \emph{Clean}, \emph{Chat} and \emph{Drink}, based on F1 scores. The class-wise F1 scores for these activities are summarized in Fig. \ref{clsf1} for all three video resolutions obtained from the two-stream and single-stream methods. From the results, we observe that the single-stream method completely fails to identify class \emph{Clean} and class \emph{Chat} (F1 scores below $10\%$) in the $64\times64$ resolution videos whereas the proposed method is still able to identify the activities based on the color-coded masks. From the figure it is also evident that as the frame resolution increases the performance of the activity classifier also increases.


\subsection{Effect of stream-wise attention}
\begin{figure}[t]
    \centering
    \includegraphics[width=\linewidth, trim={20 0 75 60},clip]{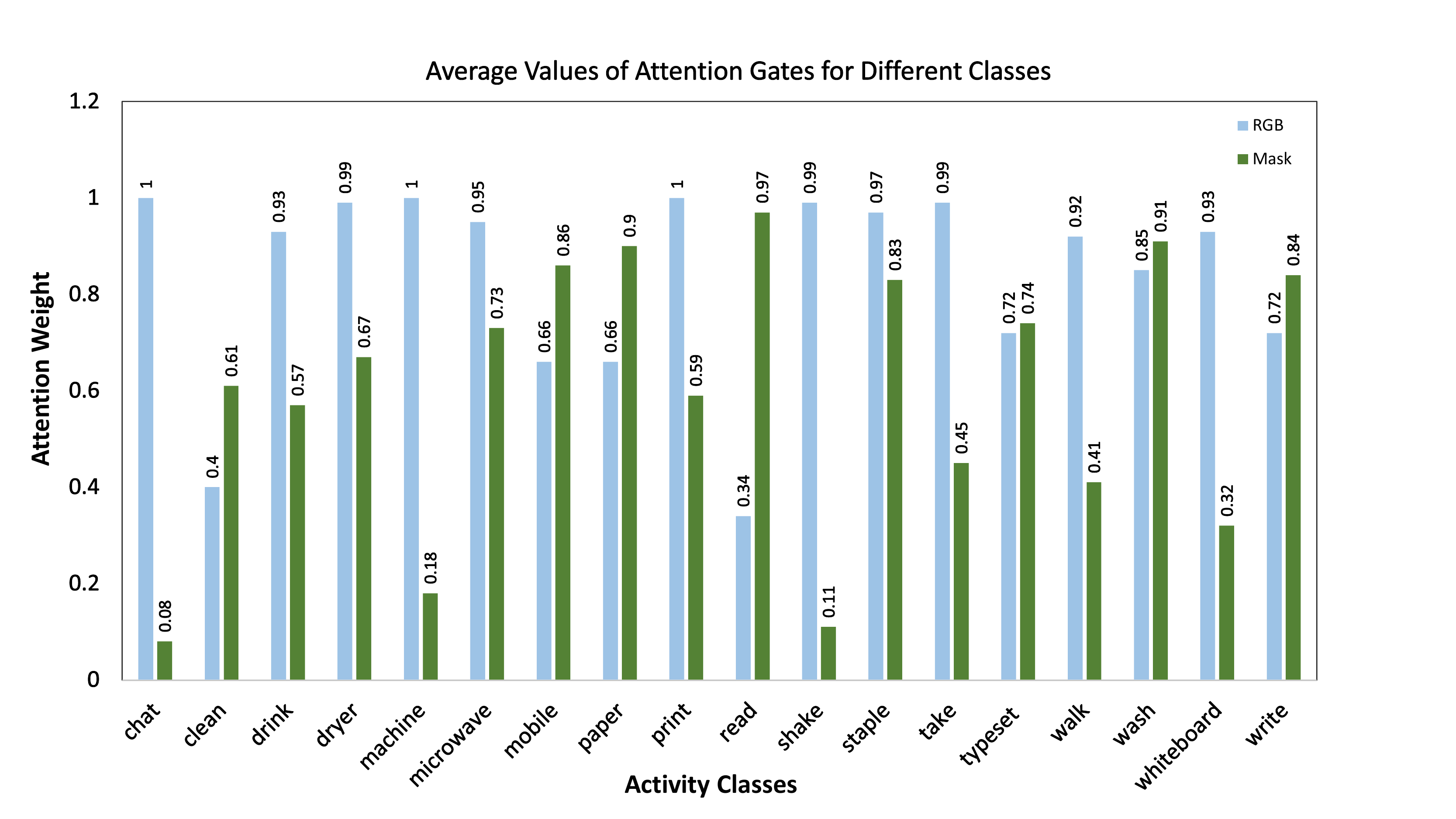}
    \caption{Average stream-wise attention weights for different classes obtained from the RGB single-stream (blue) and the segmentation mask stream (green) in the proposed architectures. We note that some classes (e.g., chat, shake) do not depend on the mask stream while other classes (e.g., mobile, paper) depend more on the color coded masked frames compared to the raw RGB frames.}
    \label{gateMean}
\end{figure}

To evaluate the impact of the stream-wise attention module, we performed experiments on the proposed system with and without this module. The results are summarized in Table \ref{StreamGateAb} for input video frame dimensions of $224\times224$. Here, we observe that using the stream-wise attention provides an absolute improvement in mean F1 score and accuracy by $2.14\%$ and $0.98\%$, respectively. The average stream-wise attention values for different classes impact of the stream gate while making predictions is demonstrated in Fig. \ref{gateMean}.

\begin{table}[t]
\caption{Effectiveness of the stream-wise attention module in activity classification performance}
\label{StreamGateAb}
\centering
\begin{tabular}{ccc}
\toprule
Proposed system configuration & Avg F1 score (\%) & Accuracy (\%) \\ \midrule
With stream-wise attention & \bf 90.44 & \bf 89.68        \\
Without stream-wise attention & 88.30    & 88.70           \\ \bottomrule
\end{tabular}
\vspace{5mm}
\caption{Effectiveness of the frame-wise attention module in activity classification performance}
\label{FrameAb}
\centering
\begin{tabular}{ccc}
\toprule
Proposed system configuration & Avg F1 score (\%) & Accuracy (\%) \\ \midrule
With frame-wise attention & \bf 90.44 &  \bf 89.68 \\
Without frame-wise attention & 89.10 & 88.50 \\ \bottomrule
\end{tabular}
\end{table}

\subsection{Effect of frame-wise attention}
\begin{figure*}[t]
    \hspace{-20mm}
    \includegraphics[height=4.5in,width=8.6in]{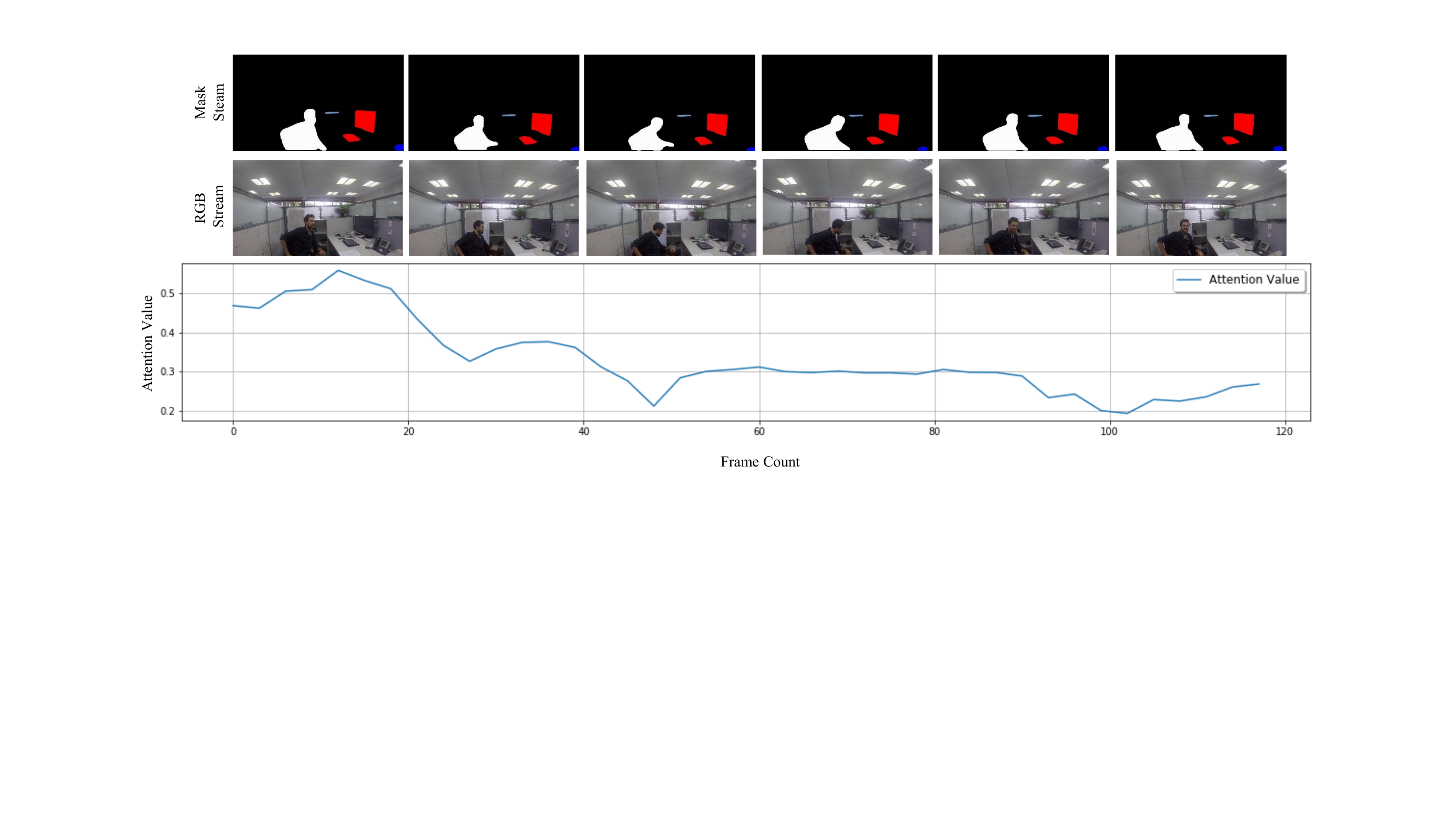}
    \vspace{-55mm}
    \caption{Variation of the frame-wise attention values for a series of sub-sampled frames obtained from a video segment containing the \emph{chat} activity class. Both the RGB and the mask frames are shown for illustration. In the mask frames, the relevant objects person, water flask and computer screen are detected and color-coded with white, blue and red, respectively.
    The frame-wise attention values increase when the person is looking towards the camera for talking.}
    \label{fwab}
\end{figure*}

To study the effectiveness of the frame-wise attention module, we ran a set of experiments including and excluding this module, while other aspects of the architecture remained the same. The results of these experiments presented in Table \ref{FrameAb} show that the frame-wise attention provides an absolute gain in mean F1 score and accuracy by $1.34\%$ and $1.18\%$, respectively. As an illustrative example, the frame-wise attention values obtained from a video sample containing the activity class \emph{chat} is shown in Fig. \ref{fwab}. We observe from this figure that the attention value is higher when the person is looking toward the camera and is more likely to be speaking to the subject (FPV camera bearer). 

\subsection{Overall results}
In the final evaluation, we compare the proposed system with the state-of-the-art I3D model and the single-stream baseline for the activity classification task. 
In this experiment, we fixed the video frame size to $224\time224$ for all methods. From the results presented in Table \ref{baseline}, we observe that the I3D method achieves an accuracy of $88.380\%$ and an F1 score of $85.647\%$ while the proposed two-stream method reaches an accuracy of $90.799\%$ and F1 score of $90.176$. Thus, the proposed model improves the performance on this task by an absolute margin of $4.209\%$ in averaged class-wise F1 score.

\begin{table}[b]
\caption{Comparison of the proposed architecture with baseline system I3D and single-stream RGB systems on videos of dimension $224\times224$.}
\label{baseline}
\centering
\begin{tabular}{lcc}
\toprule
Activity classification system & Avg F1 score (\%) & Accuracy (\%) \\ \midrule
Single-stream RGB baseline system & 89.437 &  90.669\\
I3D system  & 85.647    &  88.380          \\ 
Proposed two-stream system & \bf 90.176 & \bf 90.799\\
\bottomrule
\end{tabular}
\end{table}

\section{Discussion}

The experimental results have demonstrated that the proposed two-stream method performs superior compared to a baseline single-stream method and the state-of-the-art I3D model. The results of the proposed method are significantly better in the lower resolution videos. In this section, we intend to point out a few limitations in our experiments. Firstly, we used the ResNeXt-50 feature extractor, which is not the best nor the deepest feature extractor for this kind of task. We believe that ResNet-152\cite{he2016deep} or DenseNet-201 \cite{huang2017densely} or Wide ResNet-101 \cite{zagoruyko2016wide} may have potential for further improved performance. However, due to computational limitations and for the sake of experimental comparison between the two-stream and single-stream methods, we chose to use the smaller ResNeXt-50 network. 
Secondly, we want to note that the two-stream method performance in the high resolution videos ($224\times224$) are sub-optimal. The reason is that we were able to use a large mini-batch size while experimenting with the smaller frame sizes ($64\times64$ and $112\times112$) and we also were able to train the feature extractor in the final phase of our training. However, due to computational limitations, a similar training scheme was not possible for the two-stream method. 


We also believe that our results would improve further if we could annotate our video frames for the class-relevant object masks. In this way, our segmentation module could be fine-tuned on the FPV-O data providing improved color-coded masks. Since we had to use a pre-trained Mask R-CNN model for our segmentation module, we were not able to evaluate its performance on the FPV-O dataset. 


\section{Conclusion}
In this study, we have developed SegCodeNet, a two-stream network that uses color-coded semantic segmentation mask-based video stream in addition to the conventional RGB video stream for activity classification from wearable cameras. A pre-trained Mask R-CNN model was used to generate specific colored masks for the important objects for each action class. The feature vector from this mask stream was concatenated with the feature vector of the RGB stream and provided to a  bi-directional LSTM for activity classification. Our system provided a superior performance compared to two state-of-the-art baseline models, including a single-stream method and a hybrid optical-flow based model. 
The proposed method performs significantly better compared to the single-stream method for lower resolution videos. We have also included stream-wise and frame-wise attention modules that further improves the performance. 



\balance
\bibliographystyle{IEEEtran}
\bibliography{refs.bib}

\end{document}